\pdfoutput=1
\documentclass[runningheads]{llncs}
\usepackage[T1]{fontenc}
\usepackage[utf8]{inputenc}
\usepackage{graphicx}
\usepackage{hyperref}
\usepackage{color}

\usepackage{booktabs}
\usepackage{xcolor}

\usepackage{multirow}
\usepackage{pifont}
\newcommand{\YES}{\ding{51}}
\newcommand{\NO}{{\color{gray!50}\ding{55}}}

\usepackage[absolute]{textpos}
\begin{document}
\begin{textblock}{16}(0,0.1)\centerline{This paper was published at \textbf{TSD 2024} -- please cite the published version {\small\url{https://doi.org/10.1007/978-3-031-70563-2_22}}.}\end{textblock}
\title{Open-Source Web Service with Morphological Dictionary--Supplemented Deep Learning for Morphosyntactic Analysis of Czech}
\titlerunning{Deep Learning with Morphological Dictionary for Morphosyntactic Analysis}

\author{Milan Straka\orcidID{0000-0003-3295-5576} \and Jana Straková\orcidID{0000-0003-0075-2408}}

\authorrunning{Milan Straka \and Jana Straková}

\institute{Institute of Formal and Applied Linguistic, Faculty of Mathematics and Physics, Charles University, Malostranské nám. 25, 118 00 Prague, Czech Republic \\
\email{\{straka,strakova\}@ufal.mff.cuni.cz}}

\maketitle              %

\begin{abstract}
We present an open-source web service for Czech morphosyntactic analysis. The system combines a deep learning model with rescoring by a high-precision morphological dictionary at inference time. We show that our hybrid method surpasses two competitive baselines: While the deep learning model ensures generalization for out-of-vo\-cab\-u\-lary words and better disambiguation, an improvement over an existing morphological analyser MorphoDiTa, at the same time, the deep learning model benefits from inference-time guidance of a manually curated morphological dictionary. We achieve 50\% error reduction in lemmatization and 58\% error reduction in POS tagging over MorphoDiTa, while also offering dependency parsing. The model is trained on one of the currently largest Czech morphosyntactic corpora, the PDT-C 1.0, with the trained models available at \url{https://hdl.handle.net/11234/1-5293}. We provide the tool as a web service deployed at \url{https://lindat.mff.cuni.cz/services/udpipe/}. The source code is available at GitHub (\url{https://github.com/ufal/udpipe/tree/udpipe-2}), along with a Python client for a simple use. The documentation for the models can be found at \url{https://ufal.mff.cuni.cz/udpipe/2/models#czech_pdtc1.0_model}.

\keywords{morphosyntactic analysis \and deep learning \and morphological dictionary \and POS tagging \and lemmatization}

\end{abstract}

\section{Introduction}

Czech landscape of morphosyntactic tools widely available under favorable licensing terms and provided in an off-the-shelf manner is by no means bleak. For morphological analysis, MorphoDiTa (Morphological Dictionary and Tagger)~\cite{strakova-etal-2014-open} has been widely used in the academic circles in recent years, providing morphological analysis, morphological generation, tagging and tokenization. However, MorphoDiTa's well-known aspect is its reliance on the underlying morphological dictionary, which leads to limited performance for words not included in said dictionary, the out-of-vocabulary (OOV) words. Released in 2014, MorphoDiTa also lacks the incorporation of recent advancements such as deep learning techniques and contextualized embeddings. For both morphological and syntactic analysis, we refer to UDPipe \cite{straka-2018-udpipe}, which provides tagging, lemmatization and syntactic analysis for tens of languages, Czech included. This tool, on the other hand, depends solely on deep learning.

In this paper, we present an open-source web service and Python client for morphosyntactic analysis, which combines deep learning architecture of UDPipe~2~\cite{straka-2018-udpipe} with a rescoring by a morphological dictionary MorfFlex \cite{MorfFlexCZ20} (the core of MorphoDiTa \cite{strakova-etal-2014-open}) to enhance the effectiveness of the deep learning model. Our evaluation shows that the combined system improves over both a deep learning system and a dictionary-based system by themselves. The deep learning architecture ensures generalization for dictionary OOVs and better disambiguation, while the morphological dictionary promotes consistent outputs by disallowing invalid analyses at inference time. This leads to 50\% error reduction in lemmatization accuracy in comparison with MorphoDiTa and 35\% error reduction in lemmatization accuracy in comparison with UDPipe 2. For POS tagging accuracy, we achieve 58\% and 16\% error reduction in comparison with MorphoDiTa and UDPipe 2, respectively.

Moreover, the new model is trained on one of the largest Czech morphosyntactic resources, the PDT-C 1.0 \cite{hajic-etal-2020-prague}.

To sum up, the released tool provides segmentation, tokenization, morphological analysis, lemmatization, POS tagging and syntactic analysis. It does so by combining a deep learning model with a morphological dictionary at inference time.

\section{Related Work}

MorphoDiTa (Morphological Dictionary and Tagger) \cite{strakova-etal-2014-open} is an open-source tool for morphological analysis, which performs morphological analysis, morphological generation, tagging and tokenization of natural texts, and relies on an underlying morphological dictionary (MorfFlex \cite{MorfFlexCZ20} for Czech). MorphoDiTa uses the Czech morphological system by Jan Hajič \cite{hajic-2004}.

UDPipe \cite{straka-strakova-2017-tokenizing,straka-2018-udpipe} is an open-source tool for segmentation, tokenization, lemmatization, POS tagging, morphological analysis, and dependency parsing of natural texts. UDPipe models are available for 131 datasets of 72 languages of the Universal Dependencies project, using the universal morphosyntactic tagging system of the Universal Dependencies project \cite{nivre-etal-2020-universal}. 

Majka \cite{smerk-2007-majka,smerk-2009-majka}, with its free version Fajka, is a morphological analyser, which assigns a lemma and all possible grammatical tags to each word form on the input. Majka is available for 15 languages. Czech Majka uses a Czech morphological tagset by Jakubíček et al.~\cite{jakubicek-2011-czech}.

In this work, we combine the Czech morphological dictionary MorphoDiTa with UDPipe 2 trained on the Prague Dependency Treebank -- Consolidated 1.0 (PDT-C 1.0, \cite{hajic-etal-2020-prague}). For details on the data and the morphological tagset used, see the following Section~\ref{sec:data}.

\section{Data}
\label{sec:data}

Our model is trained on The Prague Dependency Treebank – Consolidated 1.0 (PDT-C 1.0, \cite{hajic-etal-2020-prague}), which has been recently released, and is, to our knowledge, one of the largest manually annotated Czech morphosyntactic resources.\footnote{\url{https://ufal.mff.cuni.cz/pdt-c}} The project includes and consolidates several existing Czech corpora, giving rise to the following sections:

\begin{itemize}
    \item {\sc PDT}: Prague Dependency Treebank 3.5, written texts,
    \item {\sc PCEDT}: Czech part of Prague Czech-English Dependency Treebank 2.0 and Coref 2.0,
    \item {\sc PDTSC}: Prague Dependency Treebank of Spoken Czech 2.0, spoken data,
    \item {\sc FAUST}: PDT-Faust, user-generated texts.
\end{itemize} 

The morphological layer (m-layer) of the PDT-C 1.0 is manually annotated, containing nearly 4M words (m-forms). PDT-C 1.0 uses the PDT-C tag set \cite{pdtc-manual}\footnote{\url{https://ufal.mff.cuni.cz/pdt-c/publications/Appendix\_M\_Tags\_2020.pdf}} from MorfFlex \cite{MorfFlexCZ20}, which is an evolution of the original PDT tag set devised by Jan Hajič \cite{hajic-2004}.

The surface syntax layer (analytical, a-layer) is manually annotated only partially in the 1.0 version, specifically in a part of the PDT section only, and is planned for full manual annotation in the next released version. The PDT-C 1.0 employs dependency relations from the PDT analytical level \cite{pdt-manual-a-layer}.\footnote{\url{https://ufal.mff.cuni.cz/pdt-c/publications/Appendix\_A\_Tags\_2020.pdf}}

\section{Methods}

\begin{figure}[p]
\centering
  \includegraphics[width=.8\hsize]{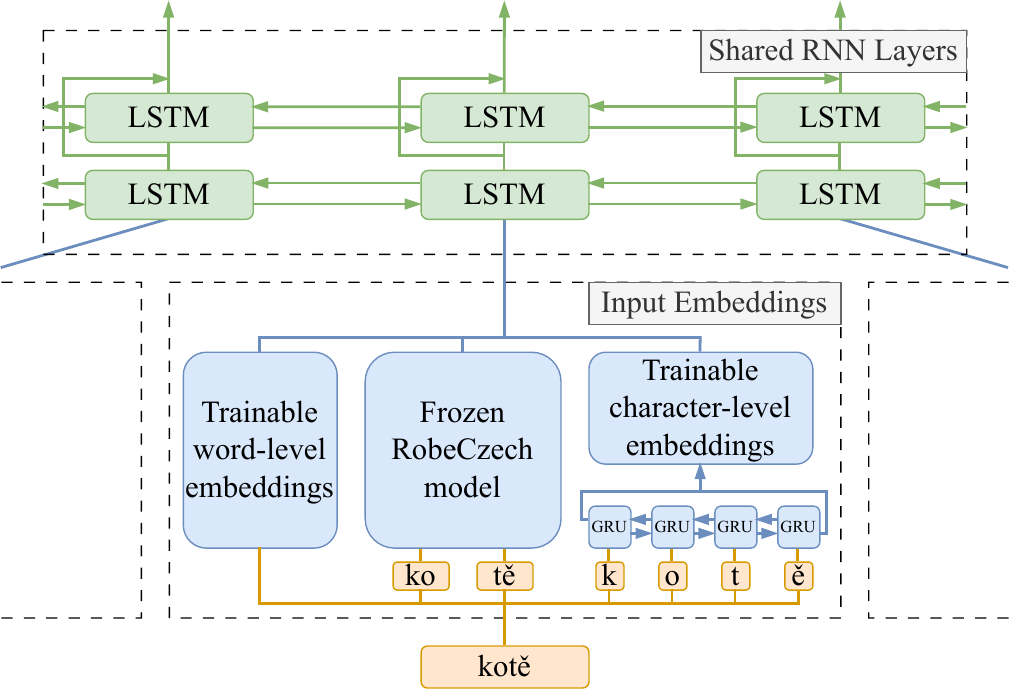}\\
  \textbf{A.} Input embeddings and shared RNN layers.
  \vspace{1em}
  
  \includegraphics[width=.8\hsize]{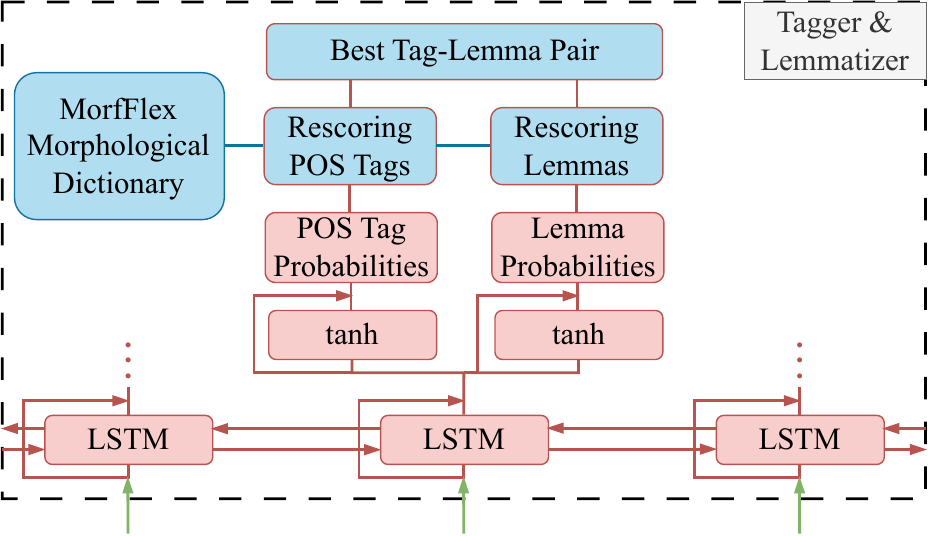}\\
  \textbf{B.} Tagger and lemmatizer.
  \vspace{1em}
  
  \includegraphics[width=.8\hsize]{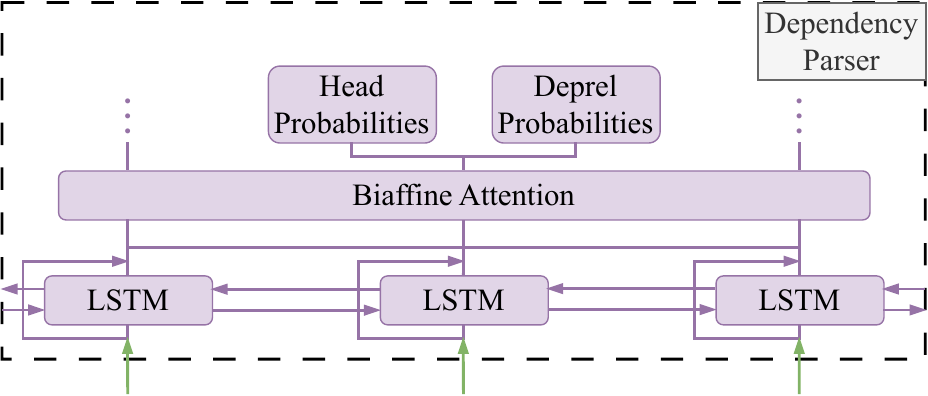}\\
  \textbf{C.} Dependency parser.
  
  \caption{An illustration of the proposed model architecture.}
  \label{fig:architecture}
\end{figure}

Our architecture is a deep learning model jointly learning morphosyntactic analysis and dependency parsing, with additional rescoring of the morphological outputs by the morphological dictionary MorfFlex \cite{MorfFlexCZ20}. The deep learning architecture is identical to the architecture of UDPipe 2 \cite{straka-2018-udpipe}, with RobeCzech \cite{straka-2021-robeczech}, a monolingual Czech pretrained language model, as a foundation. We refer to the baseline system without morphological dictionary and trained on PDT-C 1.0 as \textit{UDPipe 2} and to our system with an added morphological dictionary as \textit{Our system}. The overview of our architecture is outlined in Figure~\ref{fig:architecture}.

\subsection{Morphological Dictionary--Supplemented Deep Learning}
\label{sec:methods-dictionary}

Before we proceed to describe the employment of the morphological dictionary MorfFlex \cite{MorfFlexCZ20} at the model inference time, we first briefly describe the inference process without the presence of a morphological dictionary.

When performing POS tagging, the input to the deep learning model is a surface word form and the output is a POS tag. The model predicts a probability distribution over all POS tags, and the most probable POS tag is selected as the output. To perform lemmatization, the very same instance of the model produces also a second output, a character edit rule to convert the word form to the output lemma.\footnote{On construction of edit rules, consult \cite{straka-2018-udpipe}.} Precisely, the model again predicts a probability distribution over all edit rules as its second output, and consequently, the most probable edit rule is applied on the input form to produce an output lemma. 

Now with a morphological dictionary, though, we jointly rescore the two probability distributions, both the POS tags and the edit rule probability distributions, in the following way:

\begin{enumerate}
    \item If the input surface word form is out-of-vocabulary of the dictionary, nothing happens, and the original probability distributions predicted by the model are used without modification.

    \item If the dictionary recognizes the word form, we proceed to disambiguation via the dictionary. Most importantly, from now on, we consider only valid \hbox{form-tag}--edit rule entries found in the dictionary, disallowing any invalid \hbox{form-tag}--edit rule pairs suggested by the model. From the valid pairs, the one with the largest product of the tag probability and the edit rule probability, as predicted by the model, is selected.
\end{enumerate}

The succession of the deep learning model and the morphological dictionary ensures that an analysis is generated even for OOVs of the dictionary. Also, the deep learning model ranks the POS tags and lemmas by predicting their probability distributions for a given word form in context, using deep learning techniques and contextualized embeddings. Finally, the important contribution of the morphological dictionary is the pruning of the invalid combinations of forms, tags and lemmas.

\subsection{Re-parsing the Automatically Annotated Dependency Trees}

\looseness-1
Only a part of the PDT section of the the PDT-C 1.0 is annotated manually with dependency trees, while the other sections were annotated automatically at the time of their original creation using various automatic dependency parsers available at the time or lack syntactic annotation altogether. However, as the morphological and syntactic analyses are trained jointly in our system \cite{straka-2018-udpipe}, the heterogenous quality of various sources of automatic parsing may indirectly influence also the results of morphological analysis.

Therefore, before training the system presented here, we automatically re-parsed these sections in order to improve and level the quality of the automatic annotations. We started our system by training it on the training portion of the syntactically annotated PDT section. We then used the first version of our system to automatically parse the remaining training sections of PDT-C 1.0, and then we re-trained our system on the entire training data of the PDT-C 1.0.

\section{Results}
\label{sec:results}

\subsection{Lemmatization and POS Tagging Results}

\begin{table}[t]
    \centering
    \caption{Lemmatization accuracy [\%] on PDT-C 1.0. We consider lemmas with sense disambiguation numbers but without additional dictionary comments.}
    \label{tab:lemmatization}
    \setlength{\tabcolsep}{2.3pt}
    \begin{tabular}{lc|cccc|c|cc}
    \toprule
      \multirow{2}{*}{System} & \multirow{2}{*}{\llap{D}ictionary} & \multirow{2}{*}{\sc pdt} & \multirow{2}{*}{\sc pcedt} & \multirow{2}{*}{\sc pdtsc} & \multirow{2}{*}{\sc faust} & Macro & \multicolumn{2}{c}{Error Reduction}\\
               &         &       &       &       &       & Avg   & MorphoDita & UDPipe 2 \\
    \midrule
    MorphoDiTa & \YES    & 98.69 & 98.85 & 98.18 & 97.53 & 98.31 &            & \\ %
    UDPipe 2   & \NO     & 98.76 & 99.12 & 98.94 & 97.98 & 98.70 & 23\%       & \\
    Ours       & \YES    & 99.19 & 99.40 & 99.23 & 98.78 & 99.15 & 50\%       & 35\% \\ %
    \bottomrule
    \end{tabular}
\end{table}

\begin{table}[t]
    \centering
    \caption{POS tagging accuracy [\%] on PDT-C 1.0.}
    \label{tab:pos}
    \setlength{\tabcolsep}{2.3pt}
    \begin{tabular}{lc|cccc|c|cc}
    \toprule
      \multirow{2}{*}{System} & \multirow{2}{*}{\llap{D}ictionary} & \multirow{2}{*}{\sc pdt} & \multirow{2}{*}{\sc pcedt} & \multirow{2}{*}{\sc pdtsc} & \multirow{2}{*}{\sc faust} & Macro & \multicolumn{2}{c}{Error Reduction}\\
               &         &       &       &       &       & Avg   & MorphoDita & UDPipe 2 \\
    \midrule
    MorphoDiTa & \YES  & 96.29 & 97.00 & 96.90 & 94.87 & 96.27 &            & \\ %
    UDPipe 2   & \NO   & 98.53 & 98.64 & 98.55 & 96.84 & 98.14 & 50\% &      \\
    Ours       & \YES  & 98.78 & 98.80 & 98.77 & 97.42 & 98.44 & 58\% & 16\% \\ %
    \bottomrule
    \end{tabular}
\end{table}

Tables~\ref{tab:lemmatization} and \ref{tab:pos} display lemmatization and POS tagging results on PDT-C 1.0, respectively. Our system achieves 50\% error reduction for lemmatization accuracy and 58\% error reduction for POS tagging accuracy as compared with dictionary-based MorphoDiTa. A comparison with UDPipe 2, a deep learning model without any dictionary, shows that additional employment of a morphological dictionary at inference time reduces lemmatization error by 35\% and POS tagging error by 16\%. We provide a more detailed error analysis in Section~\ref{sec:error_analysis}.

\subsection{The Effect of Including Automatically Re-parsed Sections}

\begin{table}[t]
    \centering
    \caption{The effect of including automatically re-parsed sections on lemmatization and POS tagging accuracy [\%]. The first system is trained on the syntactically annotated part of the PDT section. The second system is trained on the entire PDT-C 1.0 with re-parsed sections. Both systems are UDPipe 2 without the morphological dictionary.}
    \label{tab:reparsing}
    \setlength{\tabcolsep}{1.9pt}
    \begin{tabular}{l|cc|cc|cc|cc|cc}
    \toprule
      \multicolumn{1}{c|}{Train}     & \multicolumn{2}{c|}{\sc pdt}   & \multicolumn{2}{c|}{\sc pcedt} & \multicolumn{2}{c|}{\sc pdtsc} & \multicolumn{2}{c|}{\sc faust} & \multicolumn{2}{c}{Macro Avg} \\
    Section          & Lemmas & POS   & Lemmas & POS   & Lemmas & POS   & Lemmas & POS   & Lemmas & POS   \\
    \midrule
    PDT   & 98.64 & 98.41 & 98.25 & 97.86 & 97.04 & 96.73 & 96.28 & 95.16 & 97.55 & 97.04 \\
    PDT-C & 98.76 & 98.53 & 99.12 & 98.64 & 98.94 & 98.55 & 97.98 & 96.84 & 98.70 & 98.14 \\
    \bottomrule
    \end{tabular}
\end{table}

In Table~\ref{tab:reparsing}, we show how the addition the automatically re-parsed sections of PDT-C 1.0 to training data on top of the syntactically annotated part of the PDT section improves the lemmatization and POS tagging accuracy. The gain is notable, with 50\% error reduction in lemmatization accuracy and 37.16\% error reduction in POS tagging accuracy. The improvement is naturally caused by the increased amount of data and domains, but the re-parsing was necessary to allow joint training of our model without deterioration on the lower-quality or missing syntactic annotations.

\subsection{Parsing Results}

For completeness, we also declare the parsing UAS and LAS score of our system, evaluated on the only section of PDT-C 1.0 with the manually annotated a-layer so far, the PDT section (see also Section~\ref{sec:data}). The system, trained on the whole PDT-C 1.0 with re-parsed treebanks, achieves UAS 94.41\% and LAS 91.48\% on the PDT section. When trained purely on the PDT section, both scores were lower by a 0.1 percent points.

\section{Error Analysis}
\label{sec:error_analysis}

\subsection{Lemmatization Improvement over UDPipe 2}
\label{sec:udpipe_lemmas_corrected}

We are now interested in error analysis between UDPipe 2 and our system, in which the added morphological dictionary rescored the model outputs at inference time. On PDT-C 1.0 test data, UDPipe 2 without dictionary made 4\,689 lemmatization errors out of all 422\,540 lemmas. Our system with a subsequent rescoring by dictionary fixed 1\,692 errors, while introducing only 143 new errors. This lead to 35\% error reduction of macro lemmatization accuracy, as shown in Table~\ref{tab:lemmatization} in Section~\ref{sec:results}.

Apparently, UDPipe 2 is willing to generate and prefer a non-existent lemma, as 49\% of its lemma errors are hallucinated non-lemmas (not in dictionary). Fortunately, it is also able to generate the correct lemma among other options, and its score is often close to the leading positions, so after the morphological dictionary prunes the most likely but invalid options, the correct lemma emerges as the winner. Consequently, the majority of dictionary-based corrections (1147, 68\%) were corrections of such fictitious lemmas.

This is particularly well observable in cases when only one valid analysis is possible according to the dictionary: See Table~\ref{tab:udpipe_analyses}, where UDPipe 2 does not achieve 100\% lemmatization and POS tagging accuracy, despite only one possible valid analysis. After correction with the dictionary, our system reached 100\% accuracy in the single analysis setting.

In more detail, of all the dictionary-fixed errors, be it lemmas or non-lemmas, 759 lemmatization errors (45\%) were sense corrections --- the lemma was correctly generated, but the lemma sense needed to be disambiguated by the morphological dictionary (``ještě-1'' $\rightarrow$ ``ještě-2'', ``Lincoln-2'' $\rightarrow$ ``Lincoln-3'', ``jak-3'' $\rightarrow$ ``jak-2'', etc.). In a few other cases (169 errors, which equals to 10\% errors), the lemma was also almost correctly generated, but the casing had to be corrected to fit the actual dictionary entry (``lovochemie'' $\rightarrow$ ``Lovochemie'', ``Fytoplankton'' $\rightarrow$ ``fytoplankton'', ``Kozoroh'' $\rightarrow$ ``kozoroh''). The remaining dictionary-fixed 764 errors (45\%) were completely incorrectly generated lemmas. We print the most frequent dictionary corrections of the completely incorrect lemmas (without lemma senses and capitalization errors) in Table~\ref{tab:udpipe_lemmas_corrected}.

\begin{table}[t]
    \centering
    \caption{Micro average accuracies [\%] for varying level of ambiguity, reflected in number of analyses; plus absolute differences between UDPipe 2 and our system, on PDT-C 1.0 test data.}
    \label{tab:udpipe_analyses}
    \setlength{\tabcolsep}{6pt}
    \begin{tabular}{c|r|rr|rr|rr}
    \toprule
      \multirow{2}{*}{Analyses} & \multirow{2}{*}{Weight} & \multicolumn{2}{c|}{UDPipe 2} & \multicolumn{2}{c|}{Ours} & \multicolumn{2}{c}{Abs. Delta} \\
      & & POS & Lemma & POS & Lemma & POS & Lemma \\
    \midrule
0 & 0.85\% & 91.01 & 91.71 & 91.01 & 91.71 & 0.00 & 0.00 \\     
1 & 41.14\% & 99.75 & 99.75 & 100.00 & 100.00 & 0.10 & 0.10 \\  
2 & 13.89\% & 97.82 & 98.00 & 98.16 & 98.59 & 0.05 & 0.08 \\    
3 & 11.99\% & 98.61 & 98.86 & 98.79 & 99.32 & 0.02 & 0.06 \\    
4 & 9.68\% & 98.26 & 98.55 & 98.45 & 99.07 & 0.02 & 0.05 \\     
5 & 4.99\% & 96.55 & 96.90 & 96.69 & 97.19 & 0.01 & 0.01 \\     
6 & 3.08\% & 97.01 & 97.49 & 97.22 & 98.07 & 0.01 & 0.02 \\     
7 & 1.93\% & 97.10 & 97.70 & 97.33 & 98.22 & 0.00 & 0.01 \\     
8 & 2.52\% & 97.80 & 98.81 & 97.96 & 99.26 & 0.00 & 0.01 \\     
9\rlap{+} & 9.92\% & 97.41 & 99.23 & 97.53 & 99.47 & 0.01 & 0.02 \\     

    \bottomrule
    \end{tabular}
\end{table}

\begin{table}[t]
    \centering
    \caption{Most frequent dictionary corrections of completely incorrect lemmas by UDPipe 2 on PDT-C 1.0 test data. Asterisk marks generated lemmas not in dictionary.}
    \label{tab:udpipe_lemmas_corrected}
    \setlength{\tabcolsep}{6pt}
    \begin{tabular}{lll}
    \bottomrule
    Forms by frequency & UDPipe 2 lemma & Dictionary-corrected lemma \\
    \midrule
úhlům, úhlech & úhlo$^\ast$ & úhel \\
Angl & Ang-2$^\ast$ & Anglie \\
Kan & kan$^\ast$ & Kanada \\
kateg & kateg$^\ast$ & kategorie \\
zataženo & zataženo-2 & zatáhnout \\
el & el-88$^\ast$ & elektrický \\
Nig & nig$^\ast$ & Nigérie \\
dožínky, Dožínky, dožínkách & dožínka$^\ast$ & dožínky \\
nedaleko & nedaleko & daleko-1 \\
Kristem & Krist$^\ast$ & Kristus-3 \\
mg & mgetr$^\ast$ & miligram \\
proklel & proklet$^\ast$ & proklít \\
nenesli, nenesly, Nenesla & nenést$^\ast$ & nést \\
jehož & jenž & jehož \\
Pierce, Piercem & Pierc$^\ast$ & Pierce \\
nindžové, nindžů & nindž$^\ast$ & nindža \\
g & gok$^\ast$ & gram \\
prostřednictvím & prostřednictvím & prostřednictví \\
dešťů & dešť$^\ast$ & déšť \\
studiích, studií & studie & studium \\
MW & MW$^\ast$ & megawatt \\
přímek & přímek$^\ast$ & přímka \\
Maď & maď$^\ast$ & Maďarsko \\
kpt & kpt$^\ast$ & kapitán \\
So & so-1$^\ast$ & sobota \\
Út & Út$^\ast$ & úterý \\

\bottomrule
    \end{tabular}
\end{table}

\subsection{Lemmatization Improvement over MorphoDiTa}

\begin{table}[t]
    \centering
    \caption{Micro average accuracies [\%] for varying level of ambiguity, reflected in number of analyses; plus absolute differences between MorphoDiTa and our system, on PDT-C 1.0 test data.}
    \label{tab:morphodita_analyses}
    \setlength{\tabcolsep}{6pt}
    \begin{tabular}{c|r|rr|rr|rr}
    \toprule
      \multirow{2}{*}{Analyses} & \multirow{2}{*}{Weight} & \multicolumn{2}{c|}{MorphoDiTa} & \multicolumn{2}{c|}{Ours} & \multicolumn{2}{c}{Abs. Delta} \\
      & & POS & Lemma & POS & Lemma & POS & Lemma \\
    \midrule
0 & 0.85\% & 81.38 & 84.93 & 91.01 & 91.71 & 0.08 & 0.06 \\       
1 & 41.14\% & 100.00 & 100.00 & 100.00 & 100.00 & 0.00 & 0.00 \\  
2 & 13.89\% & 95.45 & 98.00 & 98.16 & 98.59 & 0.38 & 0.08 \\      
3 & 11.99\% & 96.98 & 98.89 & 98.79 & 99.32 & 0.22 & 0.05 \\      
4 & 9.68\% & 94.98 & 98.06 & 98.45 & 99.07 & 0.34 & 0.10 \\       
5 & 4.99\% & 94.24 & 96.00 & 96.69 & 97.19 & 0.12 & 0.06 \\       
6 & 3.08\% & 91.84 & 95.22 & 97.22 & 98.07 & 0.17 & 0.09 \\       
7 & 1.93\% & 92.88 & 96.23 & 97.33 & 98.22 & 0.09 & 0.04 \\       
8 & 2.52\% & 95.45 & 98.58 & 97.96 & 99.26 & 0.06 & 0.02 \\       
9\rlap{+} & 9.92\% & 90.17 & 99.07 & 97.53 & 99.47 & 0.73 & 0.04 \\      
    \bottomrule
    \end{tabular}
\end{table}

The error reduction between MorphoDiTa and our system is 50\% in lemmatization macro accuracy and 58\% in POS tagging macro accuracy, as shown in Table~\ref{tab:lemmatization} and Table~\ref{tab:pos}, respectively. MorphoDiTa makes 5\,380 lemmatization errors on the PDT-C 1.0 test data, and our system made only 3\,140 lemmatization errors. Our system was correct in 3\,274 (61\%) lemmatization errors made by MorphoDiTa. But, on the other hand, our system made other, new 1\,034 errors where MorphoDiTa had the correct lemma.

A detailed analysis of errors made at varying levels of word ambiguity is shown in Table~\ref{tab:morphodita_analyses}. Besides an expected improvement of error rate in the OOV condition (0 analyses), we interestingly see an increasing trend of improvement in the more ambiguous situations, where more lemmas or lemma senses need to be disambiguated (2 and more analyses, with maximum gain at 9+ analyses). The former can be explained by the neural network generating better lemmas than the MorphoDiTa guesser in the OOV condition, as our system corrected 70\% of these errors from MorphoDita. The latter shows better disambiguation on our part. 

Indeed, the qualitative analysis of both outputs confirms that MorphoDiTa does assign some lemma from the dictionary (unlike the deep learning model of UDPipe 2, which is perfectly content with generating a hallucinated lemma), but struggles with selecting the correct lemma and/or disambiguation of lemma senses. In this, our system based on UDPipe 2 has several fundamental advantages: (i) vastly larger capacity of the neural network, (ii) pre-training of the language model on large data, and (iii) better contextualization, because it uses the contextualized word embeddings produced by a BERT-like \cite{devlin-etal-2019-bert}, Transformer-based \cite{vaswani-2017-attention} Czech model RobeCzech \cite{straka-2021-robeczech}, and these embeddings are further contextualized with a bidirectional RNN \cite{graves-2005-rnn}. MorphoDiTa, on the other hand, is an older tagger implemented as supervised, rich feature averaged perceptron \cite{collins02} with the Viterbi algorithm \cite{Forney-1973-viterbi}.

\section{Limitations}

The increased capacity and contextualization, and consequently, improved disambiguation and accuracy, come at a price in terms of computational demand and efficiency: MorphoDiTa's throughput is 10-200K words per second,\footnote{Source: \url{https://ufal.mff.cuni.cz/morphodita}} while the throughput of UDPipe 2 is 60 words per second using 1 CPU thread, or 300 words per second using 8 CPU threads, or 2k words per second on a GPU.\footnote{Source: \url{https://ufal.mff.cuni.cz/udpipe}} In conclusion, the selection of a tradeoff between efficiency and effectiveness is a consideration essential for the given task.

\section{Conclusions}

We presented an open-source web service and tool for morphosyntactic analysis. It combines a deep learning model with additional rescoring by a morphological dictionary at inference time.

In comparison with dictionary-based MorphoDiTa, we achieved 50\% error reduction in lemmatization and 58\% error reduction in POS tagging. By employing a morphological dictionary at inference time, we observed lemmatization error reduction by 35\% and POS tagging error reduction by 16\%, as compared with the deep-learning-only model UDPipe 2.

The model is deployed at \url{https://lindat.mff.cuni.cz/services/udpipe/}, the source code along with a Python client at \url{https://github.com/ufal/udpipe/tree/udpipe-2}, and the trained models at \url{https://hdl.handle.net/11234/1-5293}, under MPL 2.0 license for the source code and CC BY-NC-SA 4.0 license for the models. The documentation for the models can be found at \url{https://ufal.mff.cuni.cz/udpipe/2/models#czech_pdtc1.0_model}.

\subsubsection*{Acknowledgements}

This work has been supported by the Grant Agency of the Czech Republic under the EXPRO program as project “LUSyD” (project No.\ GX20-16819X). The work described herein has also been using data provided by the LINDAT/CLARIAH-CZ Research Infrastructure (\url{https://lindat.cz}), supported by the Ministry of Education, Youth and Sports of the Czech Republic (Project No. LM2023062).

We are grateful to our three knowledgeable and competent reviewers for their helpful comments and writing suggestions.

\bibliographystyle{splncs04}
\bibliography{TSD2024}

\begin{thebibliography}{10}
\providecommand{\url}[1]{\texttt{#1}}
\providecommand{\urlprefix}{URL }
\providecommand{\doi}[1]{https://doi.org/#1}

\bibitem{collins02}
Collins, M.: {Discriminative {T}raining {M}ethods for {H}idden {M}arkov
  {M}odels: {T}heory and {E}xperiments with {P}erceptron {A}lgorithms}. In:
  Proceedings of the 2002 Conference on Empirical Methods in Natural Language
  Processing. pp.~1--8. Association for Computational Linguistics (July 2002).
  \doi{10.3115/1118693.1118694}, \url{http://www.aclweb.org/anthology/W02-1001}

\bibitem{devlin-etal-2019-bert}
Devlin, J., Chang, M.W., Lee, K., Toutanova, K.: {{BERT}: Pre-training of Deep
  Bidirectional Transformers for Language Understanding}. In: Burstein, J.,
  Doran, C., Solorio, T. (eds.) Proceedings of the 2019 Conference of the North
  {A}merican Chapter of the Association for Computational Linguistics: Human
  Language Technologies, Volume 1 (Long and Short Papers). pp. 4171--4186.
  Association for Computational Linguistics, Minneapolis, Minnesota (Jun 2019).
  \doi{10.18653/v1/N19-1423}, \url{https://aclanthology.org/N19-1423}

\bibitem{Forney-1973-viterbi}
Forney, G.D.: {The Viterbi algorithm}. Proceedings of the IEEE  \textbf{61}(3),
   268--278 (March 1973). \doi{doi:10.1109/PROC.1973.9030}

\bibitem{graves-2005-rnn}
Graves, A., Schmidhuber, J.: {Framewise phoneme classification with
  bidirectional LSTM networks}. In: Proceedings. 2005 IEEE International Joint
  Conference on Neural Networks, 2005. vol.~4, pp. 2047--2052 vol. 4 (2005).
  \doi{10.1109/IJCNN.2005.1556215}

\bibitem{hajic-etal-2020-prague}
Haji{\v{c}}, J., Bej{\v{c}}ek, E., Hlav\'{a}\v{c}ov\'{a}, J., Mikulov{\'a}, M.,
  Straka, M., {\v{S}}t{\v{e}}p{\'a}nek, J., {\v{S}}t{\v{e}}p{\'a}nkov{\'a}, B.:
  {{P}rague Dependency Treebank - Consolidated 1.0}. In: Calzolari, N.,
  B{\'e}chet, F., Blache, P., Choukri, K., Cieri, C., Declerck, T., Goggi, S.,
  Isahara, H., Maegaard, B., Mariani, J., Mazo, H., Moreno, A., Odijk, J.,
  Piperidis, S. (eds.) Proceedings of the Twelfth Language Resources and
  Evaluation Conference. pp. 5208--5218. European Language Resources
  Association, Marseille, France (May 2020),
  \url{https://aclanthology.org/2020.lrec-1.641}

\bibitem{hajic-2004}
Haji\v{c}, J.: {Disambiguation of Rich Inflection (Computational Morphology of
  Czech)}. Linguistic Data Consortium, University of Pennsylvania (2004)

\bibitem{MorfFlexCZ20}
Haji\v{c}, J., Hlavá\v{c}ov\'{a}, J., Mikulov\'{a}, M., Straka, M.,
  {\v{S}}t\v{e}p\'{a}nkov\'{a}, B.: {MorfFlex CZ 2.0} (2020),
  \url{http://hdl.handle.net/11234/1-3186}, {LINDAT}/{CLARIN} digital library
  at the Institute of Formal and Applied Linguistics ({\'{U}FAL}), Faculty of
  Mathematics and Physics, Charles University

\bibitem{pdt-manual-a-layer}
Haji\v{c}, J., Panevov\'{a}, J., Bur\'{a}\v{n}ov\'{a}, E., Ure\v{s}ov\'{a}, Z.,
  B\'{e}mov\'{a}, A., K\'{a}rn\'{i}k, J., \v{S}t\v{e}p\'{a}nek, J., Pajas, P.:
  {ANNOTATIONS AT ANALYTICAL LEVEL. Instructions for annotators}. Institute of
  Formal and Applied Linguistics, Charles University (1999)

\bibitem{jakubicek-2011-czech}
Jakub\'{i}\v{c}ek, M., Kov\'{a}\v{r}, V., \v{S}merk, P.: {Czech Morphological
  Tagset Revisited}. In: Hor\'{a}k, A., Rychl\'{y}, P. (eds.) Proceedings of
  Recent Advances in Slavonic Natural Language Processing, RASLAN 2011. pp.
  29--42 (2011)

\bibitem{pdtc-manual}
Mikulov\'{a}, M., Haji\v{c}, J., Hana, J., Hanov\'{a}, H.,
  Hlav\'{a}\v{c}ov\'{a}, J., Je\v{r}\'{a}bek, E., \v{S}t\v{e}p\'{a}nkov\'{a},
  B., Hladk\'{a}, B.V., Zeman, D.: {Manual for Morphological Annotation,
  Revision for the Prague Dependency Treebank - Consolidated 2020 release}.
  Institute of Formal and Applied Linguistics, Charles University (2020)

\bibitem{nivre-etal-2020-universal}
Nivre, J., de~Marneffe, M.C., Ginter, F., Haji{\v{c}}, J., Manning, C.D.,
  Pyysalo, S., Schuster, S., Tyers, F., Zeman, D.: {{U}niversal {D}ependencies
  v2: An Evergrowing Multilingual Treebank Collection}. In: Calzolari, N.,
  B{\'e}chet, F., Blache, P., Choukri, K., Cieri, C., Declerck, T., Goggi, S.,
  Isahara, H., Maegaard, B., Mariani, J., Mazo, H., Moreno, A., Odijk, J.,
  Piperidis, S. (eds.) Proceedings of the Twelfth Language Resources and
  Evaluation Conference. pp. 4034--4043. European Language Resources
  Association, Marseille, France (May 2020),
  \url{https://aclanthology.org/2020.lrec-1.497}

\bibitem{straka-2018-udpipe}
Straka, M.: {{UDP}ipe 2.0 Prototype at {C}o{NLL} 2018 {UD} Shared Task}. In:
  Zeman, D., Haji{\v{c}}, J. (eds.) Proceedings of the {C}o{NLL} 2018 Shared
  Task: Multilingual Parsing from Raw Text to Universal Dependencies. pp.
  197--207. Association for Computational Linguistics, Brussels, Belgium (Oct
  2018). \doi{10.18653/v1/K18-2020}, \url{https://aclanthology.org/K18-2020}

\bibitem{straka-2021-robeczech}
Straka, M., N{\'a}plava, J., Strakov{\'a}, J., Samuel, D.: {RobeCzech: Czech
  RoBERTa, a Monolingual Contextualized Language Representation Model}. In:
  Ek{\v{s}}tein, K., P{\'a}rtl, F., Konop{\'i}k, M. (eds.) Text, Speech, and
  Dialogue. pp. 197--209. Springer International Publishing, Cham (2021)

\bibitem{straka-strakova-2017-tokenizing}
Straka, M., Strakov{\'a}, J.: {Tokenizing, {POS} Tagging, Lemmatizing and
  Parsing {UD} 2.0 with {UDP}ipe}. In: Haji{\v{c}}, J., Zeman, D. (eds.)
  Proceedings of the {C}o{NLL} 2017 Shared Task: Multilingual Parsing from Raw
  Text to Universal Dependencies. pp. 88--99. Association for Computational
  Linguistics, Vancouver, Canada (Aug 2017). \doi{10.18653/v1/K17-3009},
  \url{https://aclanthology.org/K17-3009}

\bibitem{strakova-etal-2014-open}
Strakov{\'a}, J., Straka, M., Haji{\v{c}}, J.: {Open-Source Tools for
  Morphology, Lemmatization, {POS} Tagging and Named Entity Recognition}. In:
  Bontcheva, K., Zhu, J. (eds.) Proceedings of 52nd Annual Meeting of the
  Association for Computational Linguistics: System Demonstrations. pp. 13--18.
  Association for Computational Linguistics, Baltimore, Maryland (Jun 2014).
  \doi{10.3115/v1/P14-5003}, \url{https://aclanthology.org/P14-5003}

\bibitem{vaswani-2017-attention}
Vaswani, A., Shazeer, N., Parmar, N., Uszkoreit, J., Jones, L., Gomez, A.N.,
  Kaiser, L., Polosukhin, I.: {Attention Is All You Need}. In: Proceedings of
  the 31st International Conference on Neural Information Processing Systems.
  pp. 6000–--6010. NIPS'17, Curran Associates Inc., Red Hook, NY, USA (2017),
  \url{https://proceedings.neurips.cc/paper_files/paper/2017/file/3f5ee243547dee91fbd053c1c4a845aa-Paper.pdf}

\bibitem{smerk-2007-majka}
\v{S}merk, P.: {Fast Morphological Analysis of Czech}. In: Proceedings of Third
  Workshop on Recent Advances in Slavonic Natural Language Processing, RASLAN
  2009. pp. 13--16 (2009)

\bibitem{smerk-2009-majka}
\v{S}merk, P., Rychl\'{y}, P.: {Majka – rychl\'{y} morfologick\'{y}
  analyzátor}. Tech. rep., Masarykova univerzita (2009),
  \url{http://nlp.fi.muni.cz/ma/}

\end{thebibliography}

\end{document}